\documentclass{article}
\pdfoutput=1
\usepackage{arxiv}

\usepackage[utf8]{inputenc} 
\usepackage[T1]{fontenc}    
\usepackage{hyperref}       
\usepackage{url}            
\usepackage{booktabs}       
\usepackage{amsfonts}       
\usepackage{nicefrac}       
\usepackage{microtype}      
\usepackage{xcolor}         
\usepackage{graphicx}
\usepackage{amsmath}

\title{Having Second Thoughts? Let’s hear it}

\author{%
  Jung H. Lee \\
  Pacific Northwest National Laboratory\\
  Seattle, WA \\
  \texttt{jung.lee@pnnl.gov} \\
  \And
  Sujith Vijayan \\
  School of Neuroscience\\
  Virginia Tech\\
  Blacksburg, VA \\
  \texttt{neuron99@vt.edu} \\
}

\begin{document}

\maketitle

\begin{abstract}
Deep learning models loosely mimic bottom-up signal pathways from low-order sensory areas to high-order cognitive areas. After training, DL models can outperform humans on some domain-specific tasks, but their decision-making process has been known to be easily disrupted. Since the human brain consists of multiple functional areas highly connected to one another and relies on intricate interplays between bottom-up and top-down (from high-order to low-order areas) processing, we hypothesize that incorporating top-down signal processing may make DL models more robust. To address this hypothesis, we propose a certification process mimicking selective attention and test if it could make DL models more robust. Our empirical evaluations suggest that this newly proposed certification can improve DL models’ accuracy and help us build safety measures to alleviate their vulnerabilities with both artificial and natural adversarial examples. 
\end{abstract}

\section{Introduction}
Deep learning (DL) has excelled in various domains including computer vision and natural language processing \cite{deep_review1,deep_review2}. As deep neural networks or DL models rely on a large amount of linear/nonlinear operations, their operations are intractable, and the `black box' nature of DL models hinders efficient diagnosis and repair when they fail. Consequently, it is not safe to deploy DL models into safety-critical domains \cite{Lipton2016, highstake3}. Earlier studies proposed two lines of approaches to address this challenge. The first line of studies developed interpretability/explainability algorithms that shed light on operating principles underlying their decision-making \cite{e23010018, molnar2022}. While the details may vary, they correlate DL models’ answers with input attributes or hidden layer representations and evaluate the importance of individual features on models' decisions \cite{LRP, GradCAM, IntegratedGradient, grad3, alain2018understanding, TCAV}. The second line of studies sought algorithms that can evaluate the robustness of DL's decision-making and enhance it \cite{drenkow2022systematic, zhu2022robustness, zheng2016improving, bai2021recent}. 

It is well documented that the brain exhibits a trade-off between accuracy and speed; see \cite{10.3389/fnins.2014.00150} for a review. When a fast decision is needed, the brain may make mistakes more frequently, but when more time is allowed, the accuracy of its decisions increases. Notably, the brain's decisions are not solely dependent on external stimuli. Rather, they rely on intricate interplays between bottom-up processes (from low-order sensory to high-order cognitive areas) and top-down processes (from high-order to low-order areas). As effective coordinations between brain areas may occur only with a sufficient amount of time, we hypothesize that the top-down process can make the brain's slow decisions more robust and thus that incorporating the top-down process into DL models can make their decisions robust. 

Here, we propose a certification process that mimics selective attention, one of the brain's high level cognitive functions. The searchlight hypothesis proposes that selective attention leads to more reliable decisions \cite{10.3389/fnins.2014.00150, audio-search}. When we hear a siren-like sound on the street, we search for objects associated with sirens such as police vehicles or ambulances. As we already know the distinct visual features of these objects, we can disregard irrelevant objects and via selective attention discover the audio source of the siren sound faster and more reliably. We note that selective attention involves two-stages of processing \cite{10.3389/fnins.2014.00150}. In the first stage, bottom-up sensory processing detects a stimuli, which is referred to as bottom-up attention. In the second stage, top-down processing directs low-order sensory areas to find certain types of information related to the stimuli to confirm or reject the presence of the stimuli initially surmised.

The first stage of selective attention can correspond to DL models' forward passes, and the second stage requires high-level prior knowledge (e.g., visual features of police vehicles), which may not be readily available for DL models. We propose that foundation models  trained on web-scale datasets may serve as proxy reservoirs of prior knowledge essential for top-down processes. We are particularly interested in foundational segmentation models, DINO \cite{caron2021emerging}, Grounding DINO \cite{liu2023grounding} and Segment Anything model (SAM) \cite{kirillov2023segany}, which can detect a wide range of objects and work with diverse prompts; see also \cite{langsam}. Specifically, we obtain an original prediction from a DL model and use it as a text prompt to identify the region of interest (ROI), which is used to confirm or reject the original prediction.

This certification process is referred to as `second-thought certification (STCert)'. First, we test if STCert can make DL models' decision process more robust. Then, we test if STCert can alleviate DL models’ vulnerabilities to adversarial examples. Our empirical evaluations suggest that STCert can enhance the robustness of DL and that STCert can warn adversarially manipulated inputs at high accuracy and natural adversarial examples 50\% of time.

\section{Approach}
\begin{figure}[t]
  \centering
  
   \includegraphics[width=0.5\linewidth]{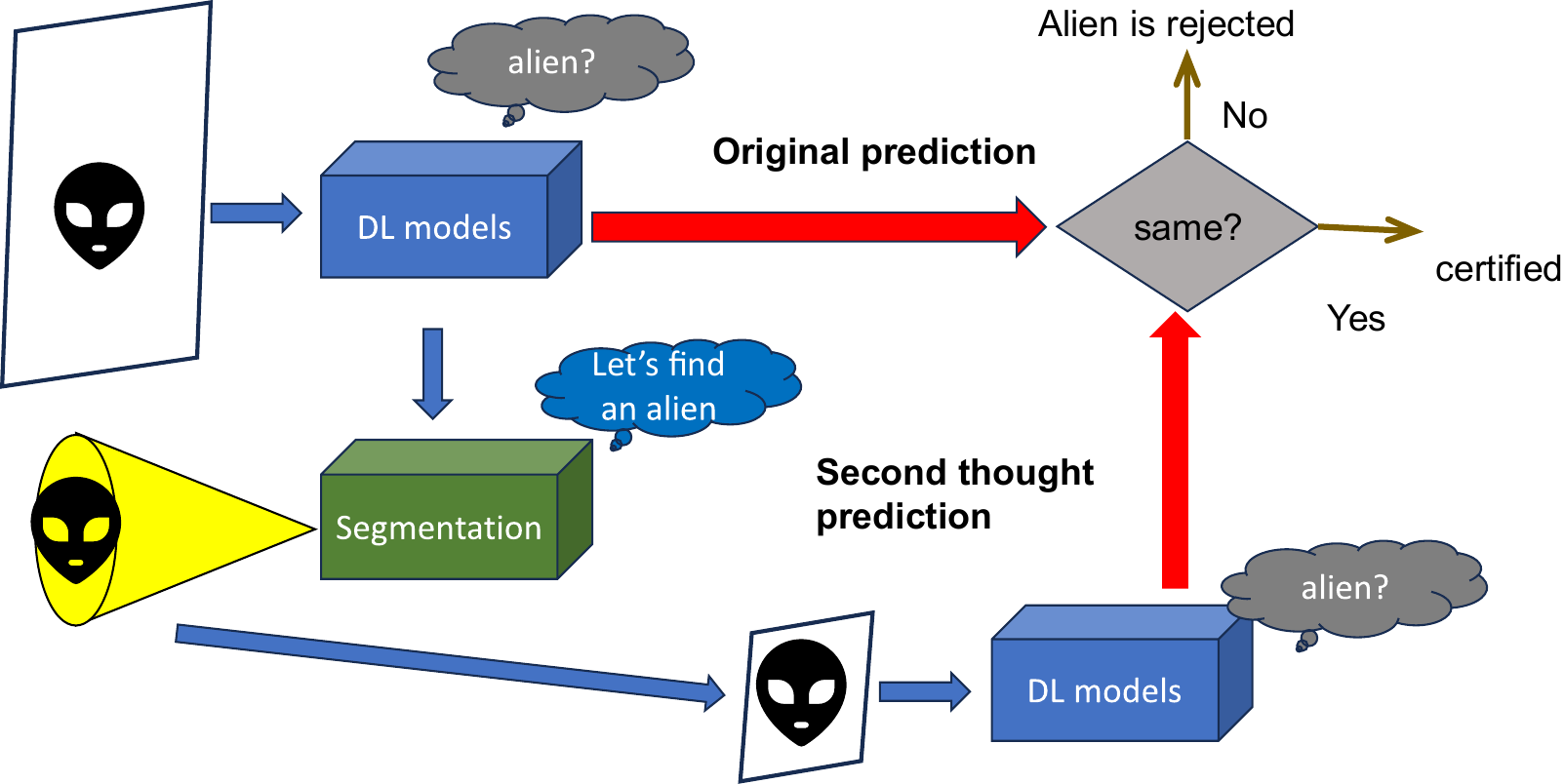}

   \caption{Workflow of STCert. In the first stage, a DL model makes an original prediction, and segmentation models (Grounding DINO and SAM in this study) identify ROI (region of interest). ROIs are used as inputs to obtain second thought predictions. Finally, STCert compares the original and second thoughts predictions to determine whether the original predictions can be certified.}
   \label{fig:fig1}
\end{figure}

STCert mimics the two stage process of selective attention (Fig. \ref{fig:fig1}). In the first stage, we obtain the original prediction from a DL model, which is an alien in the figure. In the second stage, we use two foundation models (Grounding DINO \cite{liu2023grounding} and SAM \cite{kirillov2023segany}) to detect a ROI related to the original prediction (i.e., alien in Fig. \ref{fig:fig1}) and use it as a secondary input to obtain a second thoughts prediction. In this study, we use an open-source package named `lang-SAM' \cite{langsam} to combine these two models. If a single ROI is identified, we test whether the original and second thought predictions are the same or not. If multiple ROIs are identified, we test whether the original answer is the same as one of the second thought predictions. If the original prediction is the same as or a part of second thought predictions, we consider the original answer a certified prediction. If no ROI is identified with the fine-grained class name, its super-class is used as an alternative prompt. Below, we examine two different types of STCert, which are naive STCert and context-aware STCert.
\subsection{Normal Dataset}
We evaluate STCert using Image Classifiers trained on ImageNet \cite{imagenet}. ImageNet contains examples of 1000 synsets of WordNet \cite{miller-1994-wordnet}, each of which is a part of a well-defined hierarchy. That is, each class has multiple levels of hypernyms or hyponyms. For instance, in the ImageNet, `dog' has 152 hyponyms (species) and 7 hypernyms such as `canine' and `mammal'. Differentiating between dog species is challenging to even humans, and more importantly, confusion between them is not as dangerous as confusion between a dog species and a lion. With this possibility in mind, we split errors into intra- and inter-category errors. Inter-category errors denote cases, in which predictions and ground truth labels belong to different categories. Intra-category errors denote cases, in which predictions and ground truths belong to the same categories. We consider inter-category errors and intra-category errors to be high and low risk ones, respectively, and evaluate the capability of STCert on high and low risk errors, respectively.

We note that the ImageNet hierarchy (inherited from WordNet) is not homogeneous for all classes. For instance, `German Shepherd' has 6 hypernyms, whereas `Jay' class has 8 hypernyms, making the comparison of classes’ categories difficult. To address this issue, we adopt the subsets of ImageNets curated to study DL's robustness \cite{robustness}. Specifically, we use predetermined 5 subsets, Mixed\_10, Mixed\_13, Living\_9, Big\_12 and geirhos\_16. These subsets (referred to as datasets below) contain ImageNet classes belonging to 10, 13, 9, 12 and 16 categories (a.k.a. super-classes), and we use the robustness package \cite{robustness} to select validation images from the selected classes/categories. In this study, we use validation examples included in these subsets to evaluate STCert. We do not modify or fine-tune DL models, and thus, DL models can choose one of 1,000 fine-grained classes as a prediction. We use NLTK toolkit \cite{nltk} to identify the prediction's hypernyms and determine if the prediction is hyponyms of any categories in the datasets; we ignore other hypernyms different from the categories in the database. 

\subsection{Adversarial inputs}
We further evaluate whether STCert can alleviate DL models' vulnerabilities to adversarial inputs. In this study, we consider both artificial and natural adversarial inputs. Artificial adversarial inputs are created by using autoattack \cite{autoattack}, which relies on two custom Projected Gradient Descent (PGD), FAB \cite{croce2020minimally} and Square Attack \cite{andriushchenko2020square}. We use standard autoattack with $\epsilon=0.03$ and norm=`Linf'. For natural adversarial inputs, we use ImageNet-A \cite{hendrycks2021nae}, which contains naturally occurring examples that confuse a wide range of ImageNet models.

\subsection{ImageNet models}
We evaluate STCert using a set of ImageNet models, ResNet18, ResNet50 \cite{resnet}, VGG19\_bn (with batch normalization) \cite{simonyan2015deep}, DenseNet121 \cite{huang2018densely}, ResNext101 \cite{xie2017aggregated} and vision transformer (VIT) \cite{vaswani2023attention}. All models are implemented via Timms python package \cite{rw2019timm}. It should be noted that all models are not modified or fine-tuned on the datasets (i.e., Imagenet subsets) discussed above. All experiments and analysis are conducted using a workstation with Intel core  9 CPU and RTX 4090.
 
\subsection{Naive STCert}
In the first and naive approach, we first create a mask covering ROI associated with an original prediction by combining Grounding DINO and SAM (see \cite{langsam} for more details) and use it to remove all pixels outside the identified ROI  (Eq. \ref{eq1}). Once all non-relevant pixels are removed, we use a common ImageNet preprocessing to create inputs for the second thought predictions. 
\begin{equation}\label{eq1}
    x=\vec{x}\cdot \vec{m}+(1-\vec{m})\cdot \vec{B}
\end{equation}
,where $\vec{x}$, $\vec{m}$ and $\vec{B}$ denote an image, binary mask and blank image; where $\cdot$ denotes the element-wise product.

\begin{figure}[t]
  \centering
  
   \includegraphics[width=0.9\linewidth]{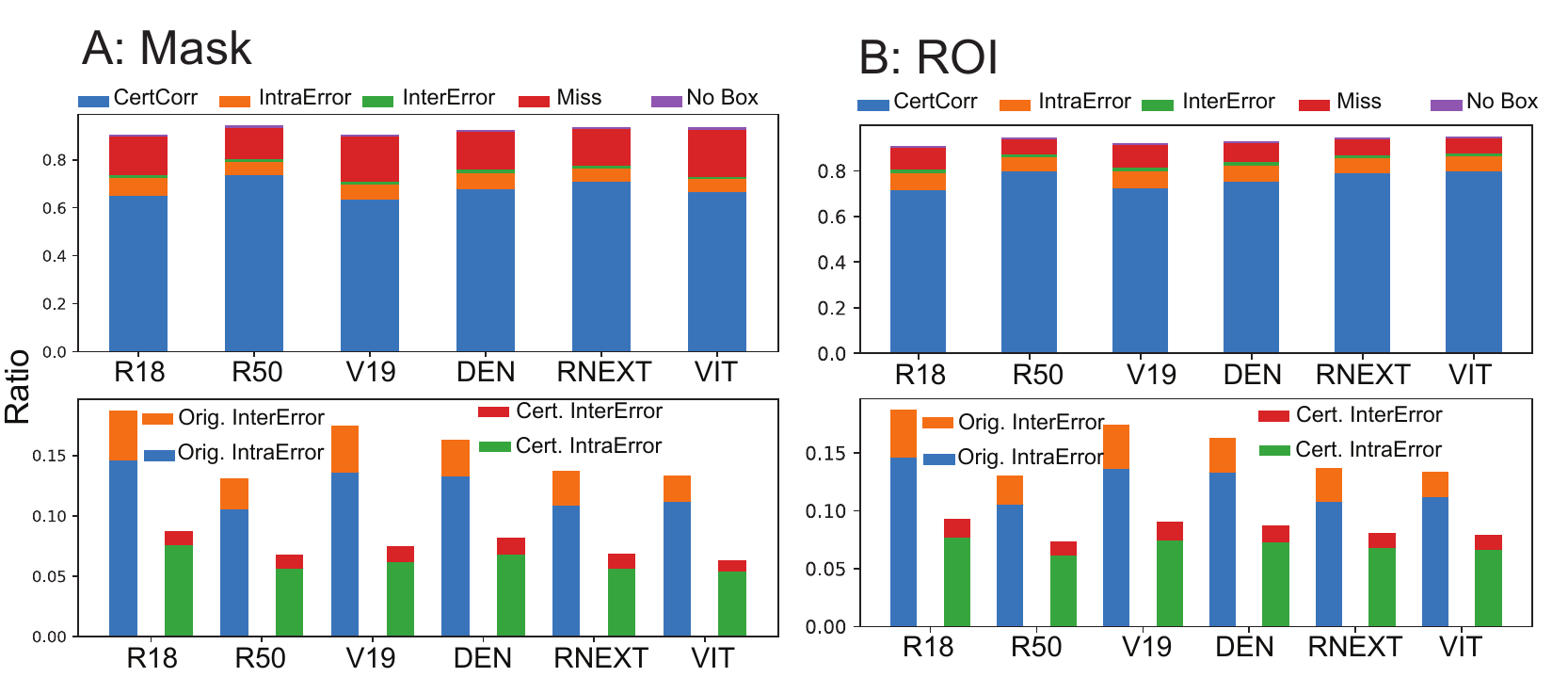}

   \caption{Certified Predictions of Mixed\_10. (A), Certified predictions on the masks of objects. (B), Certified predictions on ROI (see the text). The top row shows the ratio of certified predictions depending on the model tested. The certified answers are split into correct (CertCorr shown in blue bars), intra-category error (IntraError, shown in orange bars) and inter-category error (InterError shown in green bars). Additionally, we report how often STCert fails to certify the originally correct prediction (miss shown in red bars), and it fails to detect bounding boxes regarding the original prediction (No Box shown in purple bars); foundation segmentation models sometimes cannot detect any boxes related to given prompts. The bottom row compares Inter- and Intra-Error between original and certified predictions. Blue and orange bars denote original Intra- and Inter-Error, respectively, whereas green and red denote certified Intra- and Inter-Error, respectively. }
   \label{fig:fig2}
\end{figure}

As all non-relevant pixels are removed, the second thought predictions are determined by the pixels of ROI. We note that the original and second thought predictions of ResNet18 are consistent with 74\% of the validation images from Mixed\_10 (once again, a subset of ImageNet). This is also valid for all other datasets (Table\ref{table:1}). When the original and second thought predictions are consistent, we consider the original prediction `certified’. If not, the original prediction fails the STCert and is rejected. 
\begin{table}[]
\begin{center}
\caption{Ratio of trials, in which the original and second thought predictions are consistent. M10, M13, L9, B12 and G16 denote, Mixed\_10, Mixed\_13, Living\_9, Big\_12 and Geirhos\_16, respectively.} \label{table:1}
\begin{tabular}{lccccc}
\hline
model & M10   & M13   & L9    & B12   & G16   \\ \hline
R18   & 0.736 & 0.707 & 0.731 & 0.660 & 0.642 \\ \hline
R50   & 0.803 & 0.778 & 0.803 & 0.737 & 0.706 \\ \hline
V19   & 0.710 & 0.682 & 0.715 & 0.650 & 0.638 \\ \hline
DEN   & 0.759 & 0.731 & 0.741 & 0.686 & 0.649 \\ \hline
RNEXT & 0.776 & 0.743 & 0.773 & 0.719 & 0.681 \\ \hline
VIT   & 0.729 & 0.706 & 0.721 & 0.665 & 0.654 \\ \hline
\end{tabular}
\end{center}
\end{table}
Next, we compare certified predictions’ accuracy with that of original predictions. In the experiment, we split certified predictions into three categories, correct answers (CertCorr), inter-category error (InterError) and intra-category error (IntraError). We also observe how often STCert would reject the correct original predictions, which is indicated as `miss' trials. As shown in the top row of Fig. \ref{fig:fig2}A, the accuracy of certified predictions range from 0.65 to 0.74 depending on the models. The miss and InterError rates are much lower than IntraError. To determine whether certified predictions are more accurate and more reliable than original predictions, we compare IntraError and InterError between original and certified predictions. As shown in the bottom row of Fig. \ref{fig:fig2}A, the error rates are substantially reduced, especially in InterError. We observe the same trend in the other 4 datasets  (Supplementary Fig. \ref{fig:fig3}).

Further, we visually inspect the inter-category errors in the certified answers in Mixed\_10. In 6 cases (Supplementary Fig.\ref{fig:sup1} A-F), we find the ground truths and the certified predictions co-exist in the images. In some cases, the incorrect certified predictions and the ground truth labels are closely related. For instance, a certified prediction is `paddle', when the ground truth is a canoe (Supplementary Fig. \ref{fig:sup1}). In 1 case (Supplementary Fig. \ref{fig:sup1}H), ground truth labels are in fact incorrect. 

These results suggest that STCert can help DL models make more accurate and robust decisions. However, we note that STCert reduces overall accuracy of the models by roughly 10\% or higher. In some safety-oriented domains, such drop in accuracy may be acceptable in exchange for enhanced reliability, but for more general domains, more accurate certified prediction is desirable. To alleviate this drop in accuracy, we propose the second STCert below.

\subsection{Context-aware STCert}
In the naive approach above, all pixels unrelated to the object originally predicted are zeroed out, which makes the inputs have substantially different statistics (e.g., mean pixel values). Since DL models can learn to use pixel statistics, we assume that these zeroed out pixels may disrupt DL models’ operations too strongly and that filling them out with natural pixels may restore the accuracy. To obtain a second thought predictions determined by the original prediction only, the ideal backgrounds need to be related to the original prediction or semantically neutral. As it is not clear how to synthesize semantically neutral backgrounds, we resize ROIs to the input sizes native to the DL models ( Fig. \ref{fig:fig1}). Specifically, we infer the bounding box of the object originally predicted, crop it out and resize it to the 224-by-224 (native input size to ImageNet models). With these ROIs with additional backgrounds, we evaluate the rates of CertCorr, IntraError, InterError, Miss and No-box trials and report them in the top row of Fig. \ref{fig:fig2}B. As shown in the figure, the CertCorr rate is increased for all 6 models. We further compare InterError and IntraError between original and certified predictions and made two observations  (Bottom row of Fig. \ref{fig:fig2}B). First, both IntraError and InterError rates are reduced. Second,  InterError is reduced more significantly than IntraError. We find equivalent results in all other datasets  (Supplementary Fig. \ref{fig:fig5}).

\begin{figure*}[t]
  \centering
  
   \includegraphics[width=0.95\textwidth]{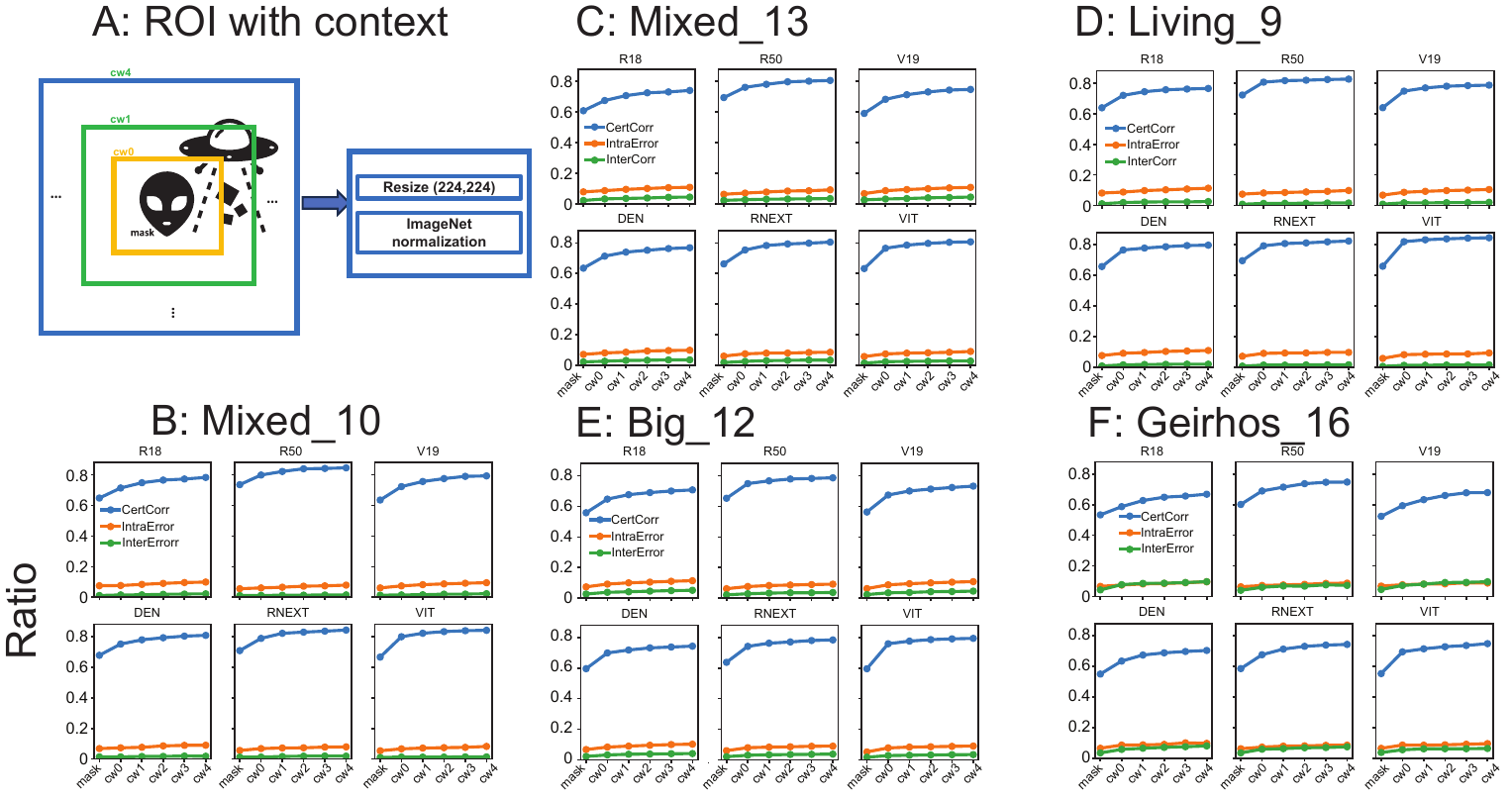}

   \caption{Evaluation of certified predictions depending on the context width (cw). (A), Schematics of our ROI selection. (B), correct certified predictions (CertCorr in blue), intra-category error (IntraError in orange) and inter-category error (InterError in green) observed on Mixed\_10.  `mask' in the $x$-axis denotes the reference point that ROIs do not contain any additional background. (C)-(F), the same as (B), but the datasets are Mixed\_13, Living\_9, Big\_12, Geirhos\_16, respectively.}
   \label{fig:fig6}
\end{figure*}

Further, inspired by the observation that DL models utilize contexts to make decisions, we gradually add more background pixels to the extracted features. As the backgrounds surrounding ROIs contain contexts closely related to original predictions, we enlarge bounding boxes. To evaluate the influence of these backgrounds (i.e., contexts), we test 5 different bounding box sizes that are referred to as context width (cw) in Fig. \ref{fig:fig6}A. As seen in Fig. \ref{fig:fig6}B, the second thought predictions' accuracy increases, as the bounding boxes grow in terms of sizes. More importantly, the growth of certified predictions’ errors remain lower than the increase of certified predictions’ accuracy, suggesting that adding backgrounds improves the accuracy of certified predictions in exchange for limited increase in errors. We repeat the same experiment with other datasets and find the equivalent results (Fig. \ref{fig:fig6}C-F).

We observe that certified predictions and ground truths are often semantically related to each other, even when they belong to other categories. We quantify this observation by estimating distances between the ground truth labels and the certified predictions using spaCy similarity measure \cite{spacy2}. Supplementary Table \ref{table:2} shows the most similar pairs of ground truth labels and certified predictions. 

\subsection{Implications for the relationships between contexts and models' decision-making}
\begin{figure*}[htb]
  \centering
  
   \includegraphics[width=0.85\textwidth]{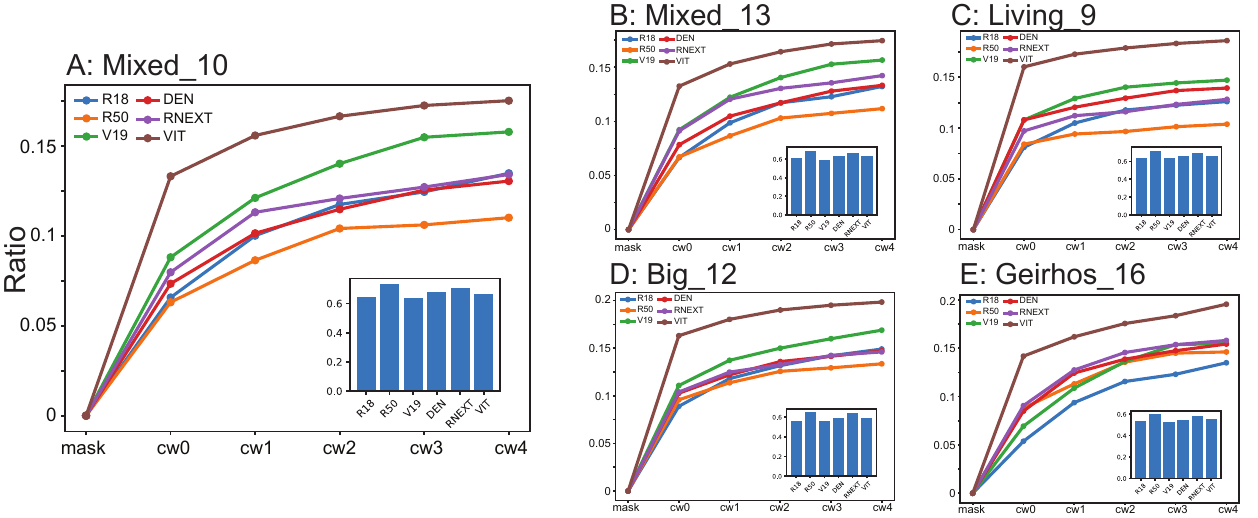}

   \caption{Accuracy increase of certified predictions depending on the context width (cw). (A), Accuracy increase observed in Mixed\_10. Individual lines show how rapidly certified predictions improve for individual models, as cw increases. `mask' in the $x$-axis denotes the reference point that ROIs do not contain any additional backgrounds. The accuracy of certified predictions with mask are shown in the inset for all 6 models.  Blue, orange, green, red, purple and brown lines denote ResNet18, ResNet50, VGG19, DenseNet, ResNext, VIT.  (B)-(E), the same as (A), but the datasets are Mixed\_13, Living\_9, Big\_12, Geirhos\_16, respectively. }
   \label{fig:fig7}
\end{figure*}

STCert uses context to make certified decisions, and the amount of context can be controlled with context width (CW, Fig. \ref{fig:fig6}A). If second thought predictions and original predictions of a DL model are always the same without context, we can interpret this result as evidence suggesting that the model does not utilize context for its decision. By contrast, if second thought predictions and original predictions become similar only when a large amount of context is provided, we can interpret it as evidence supporting that context is crucial for the model's decision. With this possibility in mind, we evaluate models' sensitivity to the context amount by evaluating the certified prediction accuracy depending on cw (Fig. \ref{fig:fig7}) and find that all models benefit from context but at different levels. VIT's accuracy increases most strongly, and ResNet50 appears to be least sensitive to the context. Interestingly, although ResNet18 and ResNet50 share the same architecture, ResNet18 requires more context to make proper decisions. Conversely, ResNet50 may utilize information encoded in ROIs (i.e., internal features) more effectively. Together with that ResNet50 has a deeper architecture, we speculate that the depth may enhance the model's capability to learn internal features.

We note that models' sensitivity to contexts depend on the identity of foreground objects (Supplementary Fig. \ref{fig:context_acc_increase}).  Interestingly, all 6 models strongly benefit from context, when the ground truth label is a `canoe'. As canoes have simple internal features, contexts may be essential for DL models to tell it from other objects. Similarly, the certified predictions' accuracy improved strongly when DL models make decisions on other objects, which have internal features common for multiple fine-grained classes under the same super-class; see Supplementary Table \ref{table:context_acc} for 10 objects, on which contexts improve the accuracy of certified predictions. These results raise the possibility that DL models may learn to use contexts when internal features are not sufficient for making proper decisions.  If we better understand what controls DL models' sensitivity to contexts (and objects' internal features), we could select the optimal architecture depending on the dataset properties related to the complexity of internal features. We believe STCert is a good starting point to develop analysis tools that evaluate DL models' sensitivity to contexts.   

\section{Adversarial Inputs}
It is well documented that DL models are vulnerable to imperceptible adversarial perturbations \cite{reviewadver1, reviewadver2}, and a more recent study \cite{hendrycks2021nae} found that they are also vulnerable to some naturally occurring examples. Certainly, both artificial and natural adversarial images pose great challenges when DL models are more widely deployed, and thus, we ask if STCert can alleviate DL models’ vulnerabilities to both artificial and natural adversarial inputs. 

\subsection{Artificial adversarial examples}
We further ask if second thought predictions are the same as original predictions to perturbed inputs. If they are different, STCert would reject original answers and tag them as unreliable predictions, which is desirable. In this study, we use `Auto Attack', one of the strongest attacks \cite{autoattack}, to create adversarial examples from all validation images in Mixed\_10.  The adversarial perturbations crafted by Auto Attack are highly effective, leading all 6 DL models to make incorrect predictions on all crafted adversarial inputs; that is, all 6 model's accuracies go down to 0\%. As we aim to test STCert’s ability to reject adversarially manipulated (incorrect) predictions, we forward the adversarially manipulated predictions and adversarial inputs to STCert and measure how many times STCert would reject or confirm predictions. When STCert rejects predictions, STCert successfully detects adversarial perturbation. When STCert confirms predictions, STCert fails to detect adversarial perturbation. We also measure the number of times STCert fails to detect the bounding boxes (ROI) of predicted objects, as this failure can also indicate original predictions that should be inspected further (i.e., successful detection of adversarial inputs).

\begin{figure}[t]
  \center
  
   \includegraphics[width=0.6\textwidth]{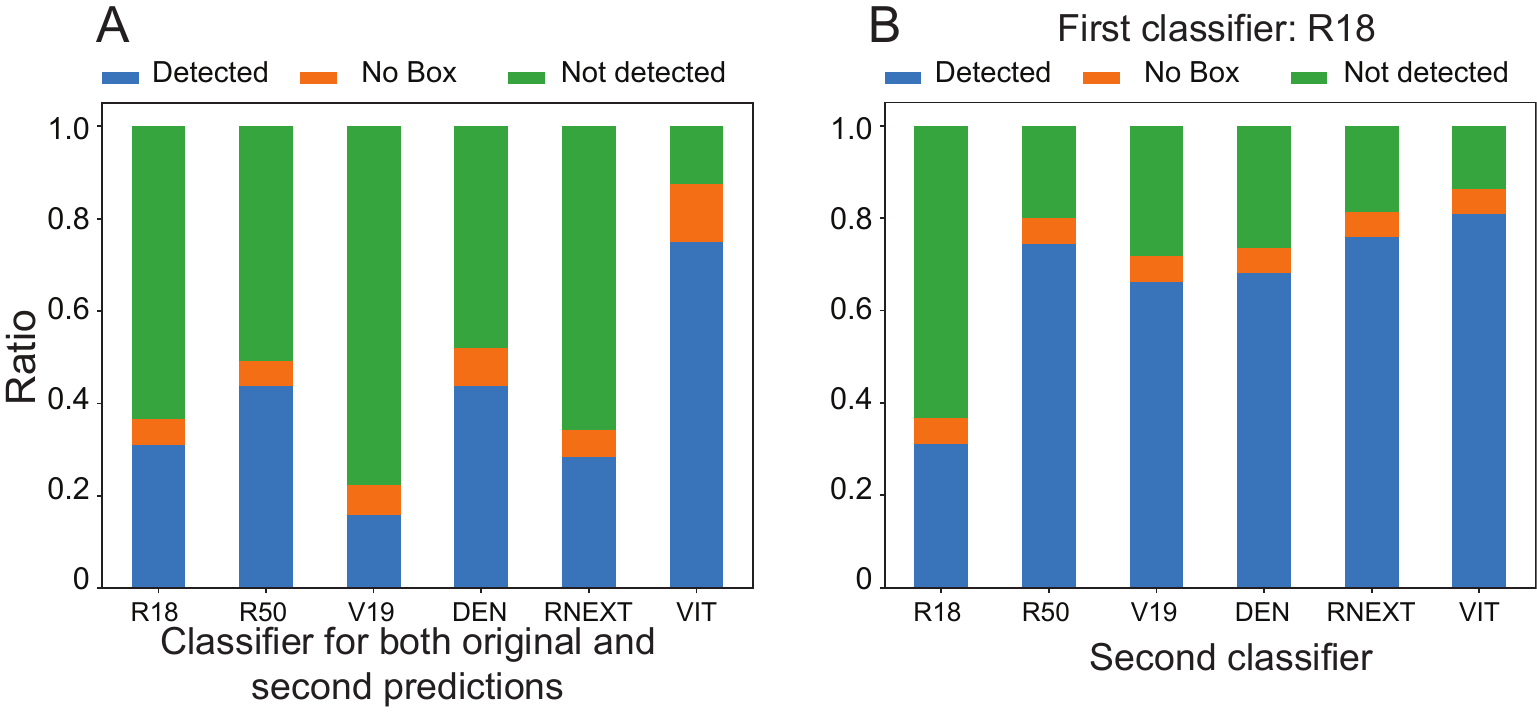}

   \caption{Detection of artificial adversarial perturbations. We ask how often incorrect predictions on artificial adversarial examples are certified by splitting the certified predictions into rejection, certification and no-box trials. If STCert rejects a prediction, it tags it as an unreliable and potentially incorrect answer. That is, it successfully detects the `weirdness' of the adversarial input. The blue bars show such detection rates. By contrast, if STCert certifies a prediction, it fails to detect the adversarial input. The green bars show the failure rates. The orange bars show how often STCert fail to detect bounding boxes related to incorrect predictions. (A), detection rates of adversarial inputs when original and second thought predictions are made by the same classifier. $x$-axis denotes the chosen classifier. (B), detection rates, when the original prediction is made by ResNet18, whereas the second prediction is made by other classifiers shown in $x$-axis.}
   \label{fig:fig8}
\end{figure}

Our results suggest that STCert detection rates strongly depend on the model (Fig. \ref{fig:fig8}A). The success rates are 22\% for VGG19 and 87.5\% for VIT. For other models, it is roughly 40-50\%. This relatively low detection rate may be explained by the fact that adversarially manipulated predictions and normal predictions often belong to the same category, and thus, the segmentation model can still find proper ROIs. With the correct ROI, the second classification can also be highly vulnerable to the adversarial inputs, since the original and second thought predictions rely on the same classifier, for which the adversarial perturbations are crafted. Thus, we can hypothesize that when original and second thought predictions are made by two different models, STCert can detect adversarial inputs more accurately. For instance, ResNet18 is used for the original prediction, but ResNet50 is used for the second thought prediction. 

To address this hypothesis, we use two different classifiers for STCert and find the detection rates to be much higher and close to 70-80\%  (Fig. \ref{fig:fig8}B), when a combination of ResNet18 and ResNet50 are used. We further test all possible combinations for 6 classifiers ($ 6 \times 6$) and measure how often STCert detects adversarial inputs and fails to detect bounding boxes. That is, we add the blue and orange bars in (Fig. \ref{fig:fig8})  to estimate the `overall success rate'. These overall success rates are 70\% or higher for all combinations  (Table \ref{tab_autoattack}), raising the possibility that STCert, which utilizes two equivalent but distinct models, can reliably detect adversarial inputs.   

\begin{table}[]
\begin{center}
        \caption{Below, we list the overall success rate of STCert on adversarial inputs, which is estimated from detection (blue bars in Fig. \ref{fig:fig8}) and no-box trials (orange bars in Fig. \ref{fig:fig8}). R18, R50, V19, D121, RN and VIT denotes ResNet18, ResNet50, VGG19\_bn, DenseNet121 and Vision Transformers, respectively.} \label{tab_autoattack}
\begin{tabular}{lllllll}
\hline
     & R18  & R50  & V19  & D121 & RN   & VIT  \\ \hline
R18  & 0.37 & 0.80 & 0.72 & 0.74 & 0.81 & 0.86 \\ \hline
R50  & 0.73 & 0.49 & 0.72 & 0.73 & 0.77 & 0.83 \\ \hline
V19  & 0.77 & 0.83 & 0.22 & 0.79 & 0.81 & 0.88 \\ \hline
D121 & 0.81 & 0.85 & 0.81 & 0.51 & 0.85 & 0.89 \\ \hline
RN   & 0.75 & 0.75 & 0.72 & 0.71 & 0.34 & 0.83 \\ \hline
VIT  & 0.91 & 0.91 & 0.91 & 0.91 & 0.92 & 0.88 \\ \hline
\end{tabular}
\end{center}

\end{table}

\subsection{Natural adversarial examples}
Notably, some naturally occurring visual patterns can be extremely difficult to detect for DL models. These natural adversarial examples may contain benign objects, but they can effectively confuse and fool a wide range of DL models, which may prevent them from being deployed in real-world applications. In our study, we ask if STCert can reject DL models’ decisions on natural adversarial examples using ImageNet-A, which includes multiple images from 200 ImageNet classes. As in artificial adversarial perturbations, we test overall success rate (i.e., combination of detection success trials and no-box trials) using all possible pairs of first and second classifiers ($6 \times 6$). Once again, the first classifier makes original predictions, and the second classifier confirms/rejects the original predictions. As shown in Table \ref{tab_imagenet-a}, the overall success rate is higher when the first and second classifiers are different.  
\begin{table}[]
\begin{center}
        \caption{Below, we list the overall success rate of STCert on imagenet-a, which is estimated from detection (blue bars in Fig. \ref{fig:fig8}) and no-box trials (orange bars in Fig. \ref{fig:fig8}).} \label{tab_imagenet-a}
\begin{tabular}{lllllll}
\hline
     & R18  & R50  & V19  & D121 & RN   & VIT  \\ \hline
R18  & 0.49 & 0.68 & 0.65 & 0.64 & 0.69 & 0.75 \\ \hline
R50  & 0.51 & 0.49 & 0.67 & 0.66 & 0.71 & 0.77 \\ \hline
V19  & 0.50 & 0.70 & 0.22 & 0.49 & 0.81 & 0.66 \\ \hline
D121 & 0.51 & 0.70 & 0.67 & 0.49 & 0.70 & 0.77 \\ \hline
RN   & 0.52 & 0.70 & 0.67 & 0.66 & 0.49 & 0.77 \\ \hline
VIT  & 0.53 & 0.71 & 0.68 & 0.67 & 0.72 & 0.49 \\ \hline
\end{tabular}
\end{center}

\end{table}
\subsection{Two classifier-based STCert on normal inputs}
\begin{figure}[t]
  \center
  
   \includegraphics[width=0.8\textwidth]{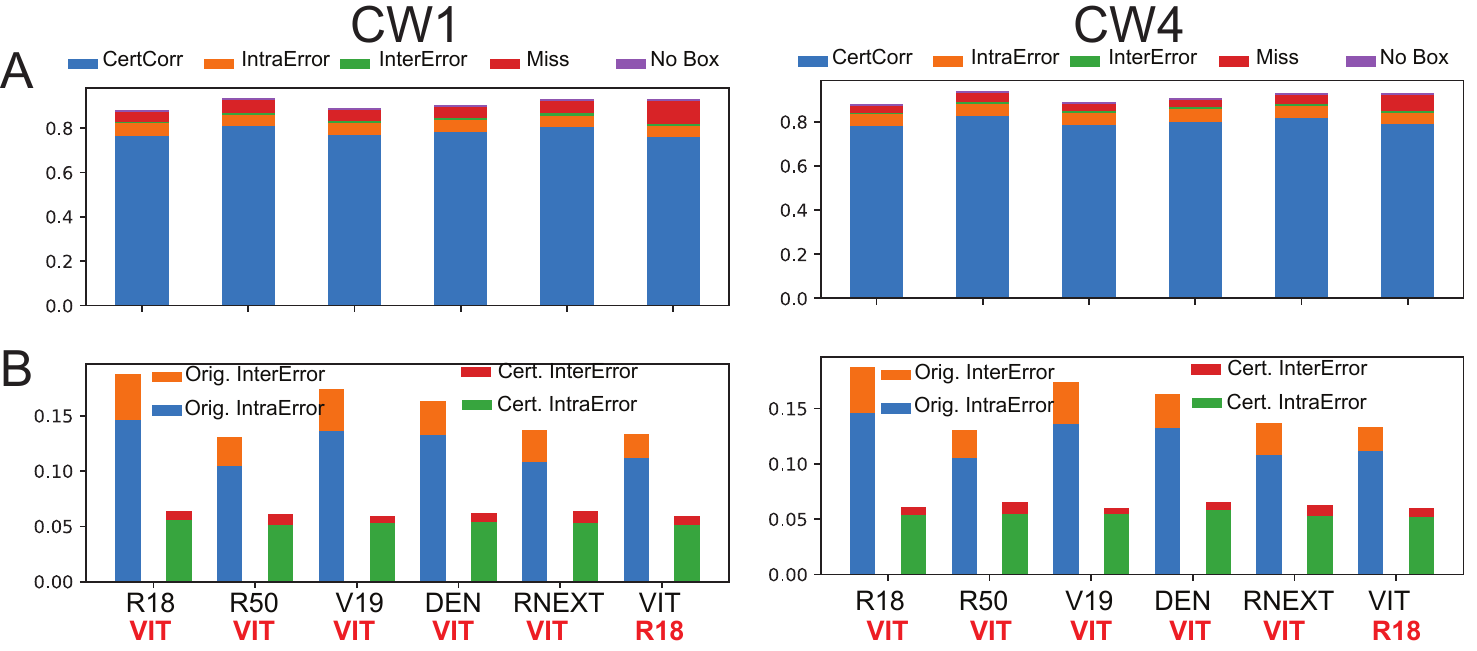}

   \caption{STCert with two classifiers. When ResNet18, ResNet50, VGG19\_bn, ResNext and DenseNet121 make original predictions, the transformer (VIT) is selected to make second thought predictions. When VIT makes the original prediction, ResNet18 is selected to make the second thought predictions. In the figure, the classifier used for the second thought prediction is shown in red at $x$-axis. (A), The ratio of certified predictions are displayed depending on the model tested. The certified answers are split into correct (CertCorr shown in blue), intra-category error (IntraError, shown in orange) and inter-category error (InterError shown in green). Additionally, we report how often STCert fails to certify the originally correct prediction (miss shown in red), and it fails to detect bounding boxes regarding the original prediction (No Box shown in purple); foundation segmentation models sometimes cannot detect any boxes related to given prompts. (B), Comparing Inter- and Intra-Error between original and certified predictions. The blue and orange bars denote original Intra- and Inter-Error, respectively, whereas green and red ones denote certified Intra- and Inter-Error, respectively. The red text denotes the classifier used to make the second prediction. The first and second column denote the results with CW1 and CW4, respectively.}
   \label{crossnormal}
\end{figure}

The results above suggest that STCert can detect incorrect predictions on artificial and natural adversarial examples more reliably when two different classifiers are used for STCert. Then, does STCert work better in general when two classifiers are used to make original and second thought predictions? To address this question, we test two classifier-based ST on normal test examples in Mixed\_10. We observe that STCert can still 1) confirm correct answers (Fig. \ref{crossnormal}), and 2) it detects incorrect answers more reliably ( Fig. \ref{fig:fig2} VS Fig. \ref{crossnormal}), when two different classifiers are used for STCert. As using two classifiers may increase the necessary cost for STCert, the two classifier-based STCert may be suitable for high-stakes and risk problems, which can justify the cost. 

\section{Discussion}

External stimuli alone do not dictate our perception. We rely on various endogenous information such as context, expectation and attention as well; see \cite{topdown} for a review. Such information is believed to have originated from higher order cognitive areas (e.g., prefrontal cortex) and be projected to sensory cortices via top-down pathways. The `searchlight hypothesis’ \cite{audio-search} suggests that selective attention may act like a filter removing non-relevant information. Notably, proper high-level information (like common sense) is needed for selective attention. Our STCert uses the foundation segmentation models as common sense to mimic selective attention. Our evaluations raised the possibility that the second thought predictions could enhance DL models’ reliability. Our findings can be summarized as follows. First, when predictions on normal inputs are certified with second thought predictions, the inter-category errors decrease significantly. Second, the most artificial adversarial examples are not certified. Third, the certification algorithms can detect roughly 50\% of the natural examples. 

Our STcert is inspired by top-down attention of the brain, and the current method is one of the simplest possible ways to mimic top-down attention. In this study, we use the same model for both original and second thought predictions to focus on testing the utility of STCert. We expect, however, a secondary model can be optimized further to process the extracted ROIs more effectively, and this custom secondary model can enhance STCert. As the extracted ROIs can be provided by the DL models trained on the original datasets, training the secondary model will not require new datasets. Further, it may be possible to leverage more contexts and knowledge to certify DL models’ predictions. For instance, by combining large language models and foundation segmentation models, we may obtain more thorough information regarding predicted objects and use it to evaluate original predictions. For instance, large language models could automatically recall that automobiles have wheels, and then we could use text prompt-based segmentation models to determine their existence in the scene. We plan to test these approaches in future studies.  
\subsection{Limitations}
Finally, we list a few major limitations of STCert. First, we underline that STCert is an error detection algorithm and does not have any means to correct incorrect decisions. Consequently,  we need additional processes/algorithms to build robust DL models (i.e., DNNs), which can reliably make correct decisions. Similarly, STCert cannot work as a common defense algorithm that enables DL models to make correct predictions/decisions even with adversarial perturbations. Second, STCert is designed for computer vision models. Extending it to other domains may be challenging because STCert requires the identification of proper contexts, and contexts are not well defined in many domains (e.g., natural language processing). Third, STCert heavily relies on segmentation models, which require large computing resources and data for training. Similarly, its inference also demands more computing power than original DL models alone. This issue may partly be mitigated by utilizing a more lightweight segmentation model trained for narrow tasks, when necessary.  Fourth, the second prediction was made using a cropped image significantly different from common (ImageNet) training images, which may reduce the accuracy of the second prediction. This issue may be mitigated by training the second classifiers on ROIs explicitly, which we plan to test in the future.

\subsection{Broader Impacts}
We propose an error detection algorithm for DL vision models. As this algorithm aims to reduce DL models’ mistakes, it could lead to building more reliable DL models. We are not aware of any potential negative impact.

\bibliography{main}

\begin{thebibliography}{10}

\bibitem{deep_review1}
Ajay Shrestha and Ausif Mahmood.
\newblock Review of deep learning algorithms and architectures.
\newblock {\em IEEE Access}, 7:53040--53065, 2019.

\bibitem{deep_review2}
Saptarshi Sengupta, Sanchita Basak, Pallabi Saikia, Sayak Paul, Vasilios Tsalavoutis, Frederick Atiah, Vadlamani Ravi, and Alan Peters.
\newblock A review of deep learning with special emphasis on architectures, applications and recent trends.
\newblock {\em Knowledge-Based Systems}, 194:105596, 2020.

\bibitem{Lipton2016}
Zachary~C. Lipton.
\newblock {The Mythos of Model Interpretability}.
\newblock In {\em ICML WHI}, 2016.

\bibitem{highstake3}
Cynthia Rudin.
\newblock {Please Stop Explaining Black Box Models for High Stakes Decisions}.
\newblock In {\em NIPS Workshop}, 2018.

\bibitem{e23010018}
Pantelis Linardatos, Vasilis Papastefanopoulos, and Sotiris Kotsiantis.
\newblock Explainable ai: A review of machine learning interpretability methods.
\newblock {\em Entropy}, 23(1), 2021.

\bibitem{molnar2022}
Christoph Molnar.
\newblock {\em Interpretable Machine Learning}.
\newblock 2 edition, 2022.

\bibitem{LRP}
Gr{\'e}goire Montavon, Alexander Binder, Sebastian Lapuschkin, Wojciech Samek, and Klaus-Robert M{\"u}ller.
\newblock {\em Layer-Wise Relevance Propagation: An Overview}, pages 193--209.
\newblock Springer International Publishing, Cham, 2019.

\bibitem{GradCAM}
Ramprasaath~R. Selvaraju, Michael Cogswell, Abhishek Das, Ramakrishna Vedantam, Devi Parikh, and Dhruv Batra.
\newblock Grad-{CAM}: Visual explanations from deep networks via gradient-based localization.
\newblock {\em International Journal of Computer Vision}, 128(2):336--359, oct 2019.

\bibitem{IntegratedGradient}
Mukund Sundararajan, Ankur Taly, and Qiqi Yan.
\newblock Axiomatic attribution for deep networks.
\newblock In Doina Precup and Yee~Whye Teh, editors, {\em Proceedings of the 34th International Conference on Machine Learning}, volume~70 of {\em Proceedings of Machine Learning Research}, pages 3319--3328. PMLR, 06--11 Aug 2017.

\bibitem{grad3}
Matthew~D. Zeiler and Rob Fergus.
\newblock Visualizing and understanding convolutional networks.
\newblock {\em CoRR}, abs/1311.2901, 2013.

\bibitem{alain2018understanding}
Guillaume Alain and Yoshua Bengio.
\newblock Understanding intermediate layers using linear classifier probes, 2018.

\bibitem{TCAV}
Been Kim, Martin Wattenberg, Justin Gilmer, Carrie Cai, James Wexler, Fernanda Viegas, and Rory sayres.
\newblock Interpretability beyond feature attribution: Quantitative testing with concept activation vectors ({TCAV}).
\newblock In Jennifer Dy and Andreas Krause, editors, {\em Proceedings of the 35th International Conference on Machine Learning}, volume~80 of {\em Proceedings of Machine Learning Research}, pages 2668--2677. PMLR, 10--15 Jul 2018.

\bibitem{drenkow2022systematic}
Nathan Drenkow, Numair Sani, Ilya Shpitser, and Mathias Unberath.
\newblock A systematic review of robustness in deep learning for computer vision: Mind the gap?, 2022.

\bibitem{zhu2022robustness}
Zhenyu Zhu, Fanghui Liu, Grigorios Chrysos, and Volkan Cevher.
\newblock Robustness in deep learning: The good (width), the bad (depth), and the ugly (initialization).
\newblock In Alice~H. Oh, Alekh Agarwal, Danielle Belgrave, and Kyunghyun Cho, editors, {\em Advances in Neural Information Processing Systems}, 2022.

\bibitem{zheng2016improving}
Stephan Zheng, Yang Song, Thomas Leung, and Ian Goodfellow.
\newblock Improving the robustness of deep neural networks via stability training, 2016.

\bibitem{bai2021recent}
Tao Bai, Jinqi Luo, Jun Zhao, Bihan Wen, and Qian Wang.
\newblock Recent advances in adversarial training for adversarial robustness, 2021.

\bibitem{10.3389/fnins.2014.00150}
Richard~P. Heitz.
\newblock The speed-accuracy tradeoff: history, physiology, methodology, and behavior.
\newblock {\em Frontiers in Neuroscience}, 8, 2014.

\bibitem{audio-search}
Jonathan~B Fritz, Mounya Elhilali, Stephen~V David, and Shihab~A Shamma.
\newblock Auditory attention--focusing the searchlight on sound.
\newblock {\em Curr Opin Neurobiol}, 17(4):437--455, 2007.

\bibitem{caron2021emerging}
Mathilde Caron, Hugo Touvron, Ishan Misra, Herv\'e J\'egou, Julien Mairal, Piotr Bojanowski, and Armand Joulin.
\newblock Emerging properties in self-supervised vision transformers.
\newblock In {\em Proceedings of the International Conference on Computer Vision (ICCV)}, 2021.

\bibitem{liu2023grounding}
Shilong Liu, Zhaoyang Zeng, Tianhe Ren, Feng Li, Hao Zhang, Jie Yang, Chunyuan Li, Jianwei Yang, Hang Su, Jun Zhu, et~al.
\newblock Grounding dino: Marrying dino with grounded pre-training for open-set object detection.
\newblock {\em arXiv preprint arXiv:2303.05499}, 2023.

\bibitem{kirillov2023segany}
Alexander Kirillov, Eric Mintun, Nikhila Ravi, Hanzi Mao, Chloe Rolland, Laura Gustafson, Tete Xiao, Spencer Whitehead, Alexander~C. Berg, Wan-Yen Lo, Piotr Doll{\'a}r, and Ross Girshick.
\newblock Segment anything.
\newblock {\em arXiv:2304.02643}, 2023.

\bibitem{langsam}
Luca Medeiros.
\newblock lang-segment-anything, 2023.

\bibitem{imagenet}
Jia Deng, Wei Dong, Richard Socher, Li-Jia Li, Kai Li, and Li~Fei-Fei.
\newblock Imagenet: A large-scale hierarchical image database.
\newblock In {\em 2009 IEEE conference on computer vision and pattern recognition}, pages 248--255. Ieee, 2009.

\bibitem{miller-1994-wordnet}
George~A. Miller.
\newblock {W}ord{N}et: A lexical database for {E}nglish.
\newblock In {\em {H}uman {L}anguage {T}echnology: Proceedings of a Workshop held at {P}lainsboro, {N}ew {J}ersey, {M}arch 8-11, 1994}, 1994.

\bibitem{robustness}
Logan Engstrom, Andrew Ilyas, Hadi Salman, Shibani Santurkar, and Dimitris Tsipras.
\newblock Robustness (python library), 2019.

\bibitem{nltk}
Edward~Loper Bird, Steven and Ewan Klein.
\newblock Natural language processing with python, 2007.

\bibitem{autoattack}
Francesco Croce and Matthias Hein.
\newblock Reliable evaluation of adversarial robustness with an ensemble of diverse parameter-free attacks.
\newblock In {\em Proceedings of the 37th International Conference on Machine Learning, {ICML} 2020, 13-18 July 2020, Virtual Event}, volume 119 of {\em Proceedings of Machine Learning Research}, pages 2206--2216. {PMLR}, 2020.

\bibitem{croce2020minimally}
Francesco Croce and Matthias Hein.
\newblock Minimally distorted adversarial examples with a fast adaptive boundary attack, 2020.

\bibitem{andriushchenko2020square}
Maksym Andriushchenko, Francesco Croce, Nicolas Flammarion, and Matthias Hein.
\newblock Square attack: a query-efficient black-box adversarial attack via random search, 2020.

\bibitem{hendrycks2021nae}
Dan Hendrycks, Kevin Zhao, Steven Basart, Jacob Steinhardt, and Dawn Song.
\newblock Natural adversarial examples.
\newblock {\em CVPR}, 2021.

\bibitem{resnet}
Kaiming He, Xiangyu Zhang, Shaoqing Ren, and Jian Sun.
\newblock Deep residual learning for image recognition, 2015.

\bibitem{simonyan2015deep}
Karen Simonyan and Andrew Zisserman.
\newblock Very deep convolutional networks for large-scale image recognition, 2015.

\bibitem{huang2018densely}
Gao Huang, Zhuang Liu, Laurens van~der Maaten, and Kilian~Q. Weinberger.
\newblock Densely connected convolutional networks, 2018.

\bibitem{xie2017aggregated}
Saining Xie, Ross Girshick, Piotr Dollár, Zhuowen Tu, and Kaiming He.
\newblock Aggregated residual transformations for deep neural networks, 2017.

\bibitem{vaswani2023attention}
Ashish Vaswani, Noam Shazeer, Niki Parmar, Jakob Uszkoreit, Llion Jones, Aidan~N. Gomez, Lukasz Kaiser, and Illia Polosukhin.
\newblock Attention is all you need, 2023.

\bibitem{rw2019timm}
Ross Wightman.
\newblock Pytorch image models.
\newblock \url{https://github.com/rwightman/pytorch-image-models}, 2019.

\bibitem{spacy2}
Matthew Honnibal and Ines Montani.
\newblock {spaCy 2}: Natural language understanding with {B}loom embeddings, convolutional neural networks and incremental parsing.
\newblock To appear, 2017.

\bibitem{reviewadver1}
Anirban Chakraborty, Manaar Alam, Vishal Dey, Anupam Chattopadhyay, and Debdeep Mukhopadhyay.
\newblock Adversarial attacks and defences: A survey, 2018.

\bibitem{reviewadver2}
Xiaoyong Yuan, Pan He, Qile Zhu, and Xiaolin Li.
\newblock Adversarial examples: Attacks and defenses for deep learning.
\newblock {\em IEEE Transactions on Neural Networks and Learning Systems}, 30(9):2805--2824, 2019.

\bibitem{topdown}
Xiao-Jing Wang.
\newblock Neurophysiological and computational principles of cortical rhythms in cognition.
\newblock {\em Physiol Rev.}, 90:1195--268, 2010.

\end{thebibliography}

\bibliographystyle{unsrt}

\clearpage
\appendix
\section{Appendix and Supplementary Material}
\setcounter{page}{1}

\setcounter{table}{0}
\renewcommand{\tablename}{Supplementary Table}

\setcounter{figure}{0}
\renewcommand{\figurename}{Supplementary Figure}

\begin{figure}[h]
  \centering
  
   \includegraphics[width=0.8\linewidth]{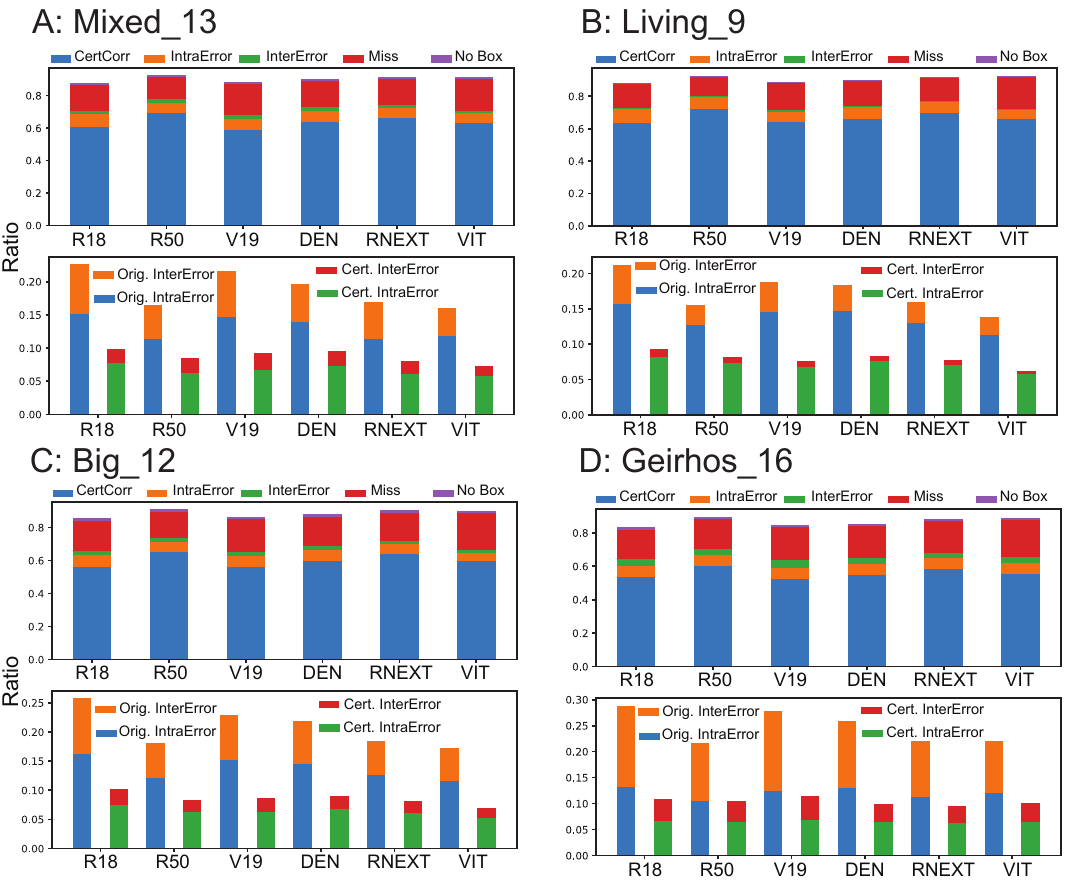}

   \caption{Comparison of certified predictions on 4 datasets. (A), the same as Fig. \ref{fig:fig2}A, but dataset is Mixed\_13. (B), the same as  Fig. \ref{fig:fig2}A, but dataset is Mixed\_13. (C), the same as  Fig. \ref{fig:fig2}A, but dataset is Big\_12. (D), the same as  Fig. \ref{fig:fig2}A, but dataset is Geirhos\_16.}  
   \label{fig:fig3}
\end{figure}

\begin{figure*}[h]
  \centering
  
   \includegraphics[width=0.9\textwidth]{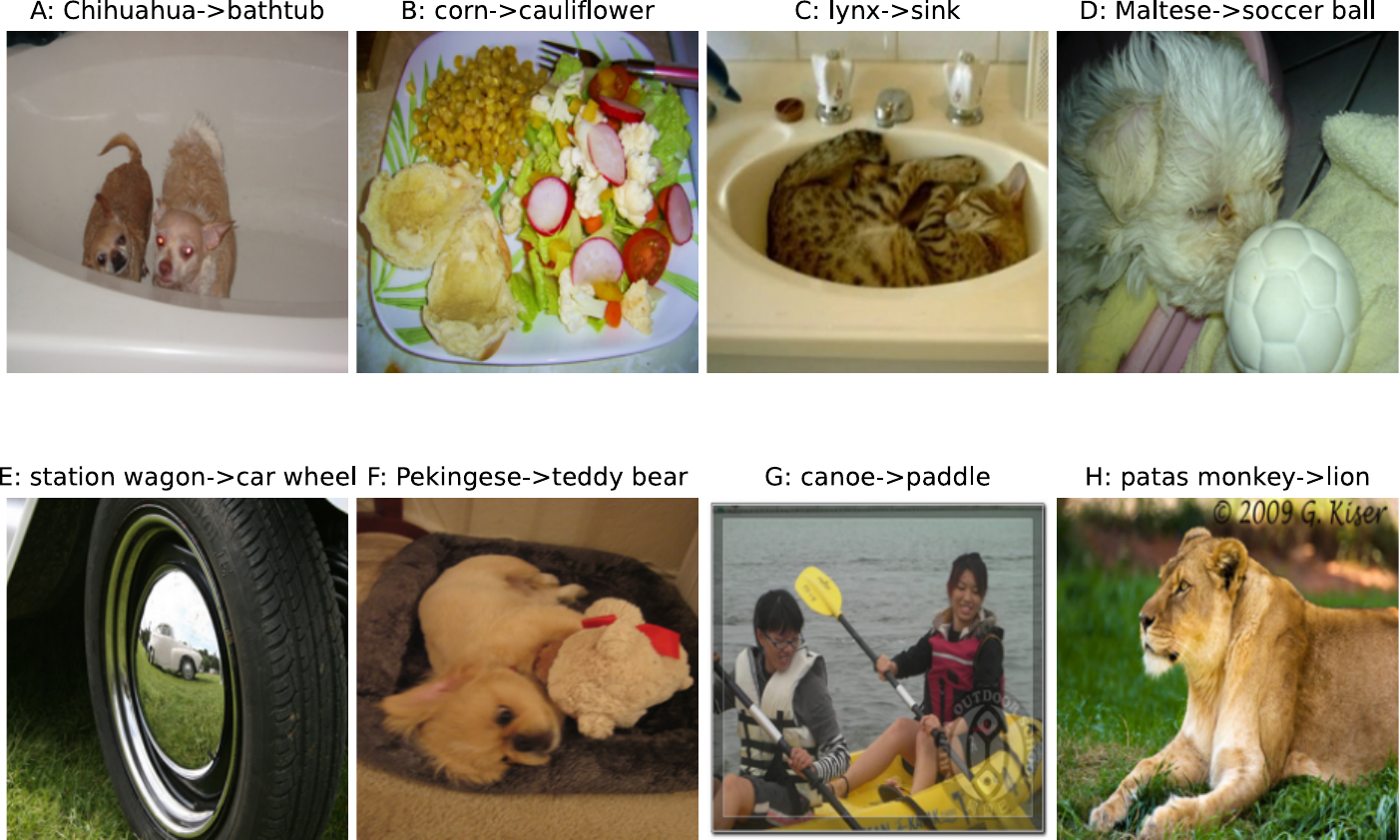}

   \caption{Examples of images that induce ResNet18 to make `IntraError', in which the ground truth labels and certified predictions belong to different categories (i.e., superclass). Ground truth labels and certified labels are shown above each image. The classes before and after the arrow denote ground truth labels and certified predictions, respectively. (A)-(F), images that include both ground truths and certified predictions. (G), image, in which ground truth and certified prediction are semantically related to each other. (H) image that contains an incorrect ground truth label.}
   \label{fig:sup1}
\end{figure*}

\begin{figure}[t]
  \centering
  
   \includegraphics[width=0.8\linewidth]{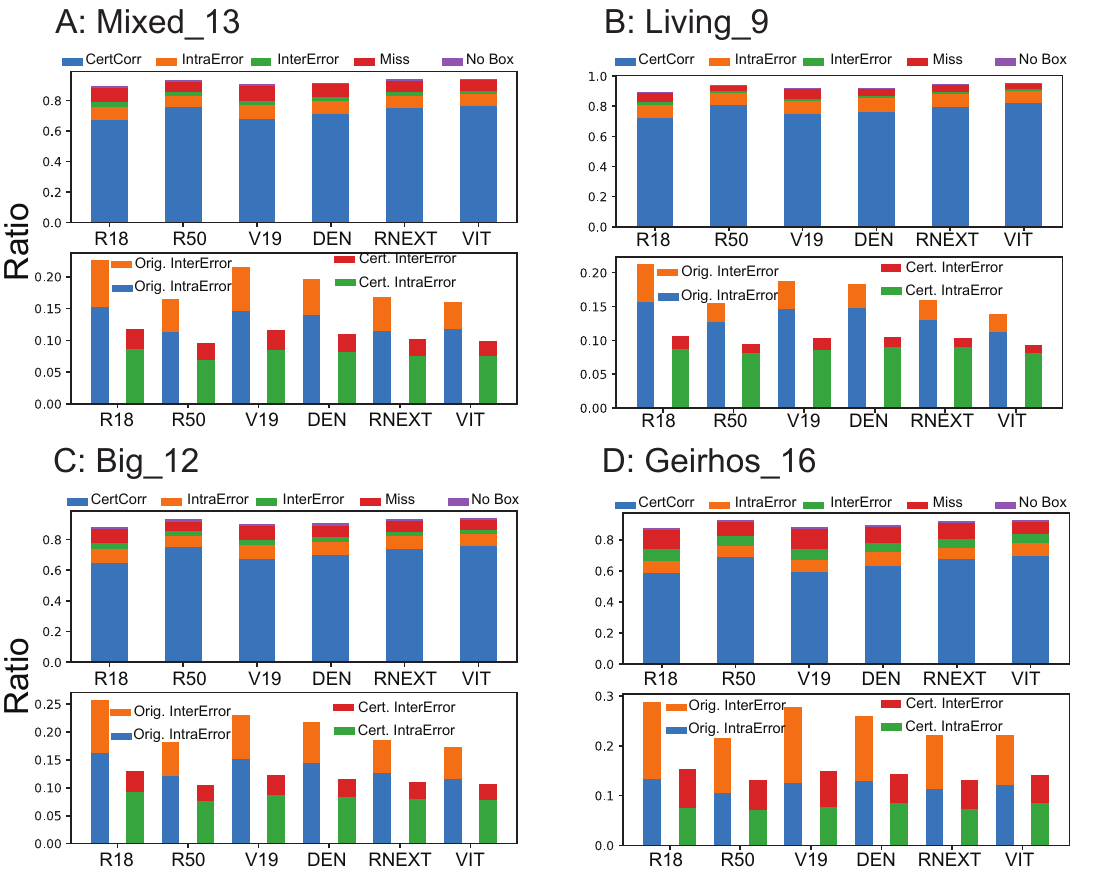}

   \caption{Comparison of certified predictions on 4 datasets. (A), the same as Fig. \ref{fig:fig2}B, but dataset is Mixed\_13. (B), the same as  Fig. \ref{fig:fig2}B, but dataset is Mixed\_13. (C), the same as  Fig. \ref{fig:fig2}B, but dataset is Big\_12. (D), the same as Fig. \ref{fig:fig2}B, but dataset is Geirhos\_16.}
   \label{fig:fig5}
\end{figure}

\begin{table}
\begin{center}
\begin{tabular}{cccc}
\hline
      & mixed\_10                                                              & mixed\_13                                                                     & living\_9                                                                  \\ \hline
top 1 & \begin{tabular}[c]{@{}c@{}}stinkhorn mushroom\\  mushroom\end{tabular} & \begin{tabular}[c]{@{}c@{}}stinkhorn mushroom\\ mushroom\end{tabular}         & \begin{tabular}[c]{@{}c@{}}tree frog\\ vine snake\end{tabular}             \\ \hline
top 2 & \begin{tabular}[c]{@{}c@{}}canoe\\ paddle\end{tabular}                 & \begin{tabular}[c]{@{}c@{}}notebook computer\\ computer keyboard\end{tabular} & \begin{tabular}[c]{@{}c@{}}banded gecko\\ tailed frog\end{tabular}         \\ \hline
top 3 & \begin{tabular}[c]{@{}c@{}}pickup truck\\ car wheel\end{tabular}       & \begin{tabular}[c]{@{}c@{}}desktop computer\\ computer mouse\end{tabular}     & \begin{tabular}[c]{@{}c@{}}loggerhead sea turtle\\ coral reef\end{tabular} \\ \hline
top 4 & \begin{tabular}[c]{@{}c@{}}corn\\ mashed potato\end{tabular}           & \begin{tabular}[c]{@{}c@{}}canoe\\ paddle\end{tabular}                        & \begin{tabular}[c]{@{}c@{}}tiger shark\\ grey whale\end{tabular}           \\ \hline
top 5 & \begin{tabular}[c]{@{}c@{}}taxicab\\ minibus\end{tabular}              & \begin{tabular}[c]{@{}c@{}}pickup truck\\ car wheel\end{tabular}              & \begin{tabular}[c]{@{}c@{}}axolotl\\ alligator lizard\end{tabular}         \\ \hline
      & big\_12                                                                & geirhos\_16                                                                   &                                                                            \\ \hline
top 1 & \begin{tabular}[c]{@{}c@{}}mushroom\\ stinkhorn mushroom\end{tabular}  & \begin{tabular}[c]{@{}c@{}}beer bottle\\ beer glass\end{tabular}              &                                                                            \\ \hline
top 2 & \begin{tabular}[c]{@{}c@{}}cliff dwelling\\ cliff\end{tabular}         & \begin{tabular}[c]{@{}c@{}}canoe\\ paddle\end{tabular}                        &                                                                            \\ \hline
top 3 & \begin{tabular}[c]{@{}c@{}}dock\\ pier\end{tabular}                    & \begin{tabular}[c]{@{}c@{}}beer bottle\\ coffee mug\end{tabular}              &                                                                            \\ \hline
top 4 & \begin{tabular}[c]{@{}c@{}}table lamp\\ lampshade\end{tabular}         & \begin{tabular}[c]{@{}c@{}}pill bottle\\ bottle cap\end{tabular}              &                                                                            \\ \hline
top 5 & \begin{tabular}[c]{@{}c@{}}drum \\ drumstick\end{tabular}              & \begin{tabular}[c]{@{}c@{}}paper knife\\ pencil case\end{tabular}             &                                                                            \\ \hline
\end{tabular}

	\end{center}
	\caption{In some IntraError examples, ground truth labels and certified predictions are semantically related with each other according to spaCy similarity measure \cite{spacy2}. This table shows 5 pairs of ground truth and certified predictions, which are strongly related with each other in terms of semantic distance.  Each column represents a dataset.}
	\label{table:2}
\end{table}

\begin{figure}[htb]
    \centering
    \includegraphics[width=0.4\linewidth]{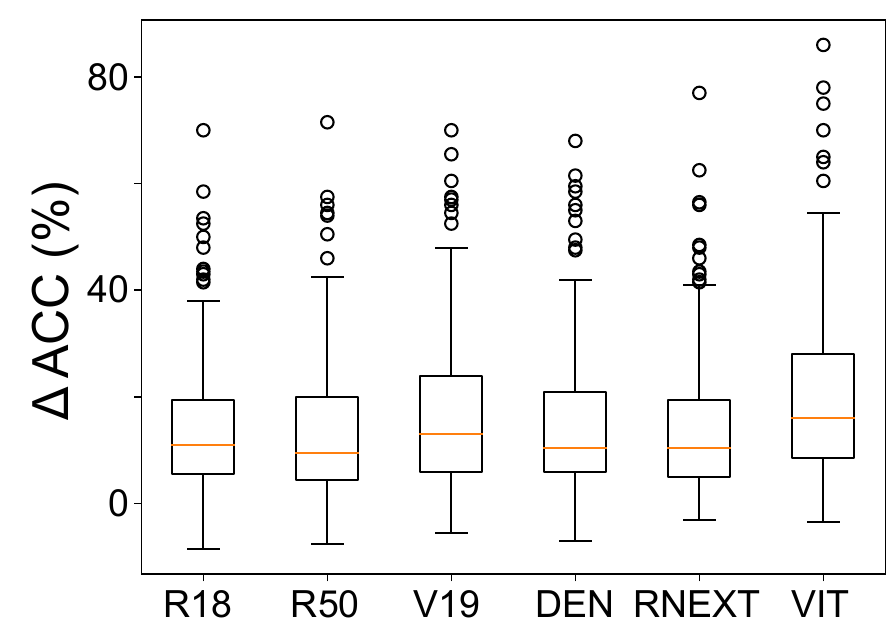}
    \caption{Accuracy increase ($\Delta$ACC) induced by contexts. We measure the difference in models' accuracy between naive STCert and context-aware STCert for all classes in 5 datasets. As we test 5 different levels of cw, we obtain 5 accuracy differences and calculate its average value. This box plot shows the distributions of these class-specific averages for 6 models. In some classes, accuracy is increased by over 50\%, and the variability of  ($\Delta$ACC) is  high, suggesting that the benefit of contexts strongly depends on the class.}
    \label{fig:context_acc_increase}
\end{figure}

\begin{table*}
	\begin{center}
		\begin{tabular}{cccc}
			\hline
			  Model& R18 & R50 & V19 \\
			\hline
			top 1 & butcher shop & butcher shop & go-kart \\
			\hline
			top 2 & howler monkey & canoe & butcher shop \\
			\hline
			top 3 & canoe & confectionery store & canoe \\
			\hline
			top 4 & cockroach & stingray & confectionery store \\
			\hline
			top 5 & dining table & mountain bike & dining table \\
			\hline
			top 6 & confectionery store & dining table & southern black widow \\
			\hline
			top 7 & go-kart & cockroach & dung beetle \\
			\hline
			top 8 & fireboat & kit fox & cockroach \\
			\hline
			top 9 & apiary & football helmet & apiary \\
			\hline
			top 10 & dung beetle & fireboat & howler monkey \\
			\hline
			  Model& DEN & RNEXT & VIT \\
			\hline
			top 1 & butcher shop & butcher shop & yellow garden spider \\
			\hline
			top 2 & canoe & canoe & canoe \\
			\hline
			top 3 & apiary & howler monkey & howler monkey \\
			\hline
			top 4 & dung beetle & confectionery store & butcher shop \\
			\hline
			top 5 & confectionery store & fireboat & cockroach \\
			\hline
			top 6 & European garden spider & football helmet & confectionery store \\
			\hline
			top 7 & howler monkey & cockroach & dung beetle \\
			\hline
			top 8 & mountain bike & southern black widow & tarantula \\
			\hline
			top 9 & cockroach & bookstore & harvestman \\
			\hline
			top 10 & fireboat & dining table & boathouse \\
			\hline
		\end{tabular}
	\end{center}
	\caption{Classes that benefited most strongly from contexts for all 6 models, ResNet18(R18), ResNet50(R50), VGG19(V19), DenseNet121(DEN), ResNext(RNEXT) and VIT. }
	\label{table:context_acc}
\end{table*}

\end{document}